\documentclass[conference]{IEEEtran}
\IEEEoverridecommandlockouts
\usepackage{cite}
\usepackage{amsmath,amssymb,amsfonts}
\usepackage{algorithmic}
\usepackage{graphicx}
\usepackage{textcomp}
\usepackage{xcolor}
\renewcommand{\figurename}{Figure}
\renewcommand{\thetable}{\arabic{table}}
\def\BibTeX{{\rm B\kern-.05em{\sc i\kern-.025em b}\kern-.08em
    T\kern-.1667em\lower.7ex\hbox{E}\kern-.125emX}}
\begin{document}

\title{Abstractive Text Summarization for Bangla Language Using NLP and Machine Learning Approaches\\

}

\author{

\IEEEauthorblockN{Asif Ahammad Miazee }
\IEEEauthorblockA{\textit{Computer Science} \\
\textit{Maharishi International University}\\
Iowa, USA \\
asifahammad7@gmail.com}
\and
\IEEEauthorblockN{
 Tonmoy Roy
}
\IEEEauthorblockA{\textit{Data Analytics \& Information Systems 
} \\
\textit{Utah State University}\\
Utah, USA \\
tonmoyroy1992@gmail.com
}
\and
\IEEEauthorblockN{Md Robiul Islam}
\IEEEauthorblockA{\textit{Computer Science} \\
\textit{William \& Mary}\\
Virginia, USA \\
robiul.cse.uu@gmail.com}

\and
\IEEEauthorblockN{Yeamin Safat}
\IEEEauthorblockA{\textit{Mymensingh Engineering College} \\
Dhaka, Bangladesh \\
yeaminsafatcp@gmail.com}

}

\maketitle

\begin{abstract}

Text summarization involves reducing extensive documents to short sentences that encapsulate the essential ideas. The goal is to create a brief summary that effectively conveys the main points of the original text. We spend a significant amount of time each day reading the newspaper to stay informed about current events both domestically and internationally. While reading newspapers enriches our knowledge, we sometimes come across unnecessary content that isn't particularly relevant to our lives. In this paper, we introduce a neural network model designed to summarize Bangla text into concise and straightforward paragraphs, aiming for greater stability and efficiency.
\end{abstract}


\section{Introduction}

Text or document summarization refers to the process of reducing a long document into concise sentences that capture the main ideas and essential information.Algorithmic summarization has become essential in our daily lives, as it reduces the effort and time required to obtain concise and relevant summaries that capture the essential information of documents. 
Summarization methods can be divided into two main types according to how they select and organize content: extractive and abstractive approaches.

Extractive summarization entails selecting the key sentences from a text using certain features and combining them to form a summary, akin to highlighting text with a highlighter. Conversely, abstractive summarization generates new sentences that encapsulate the most crucial information, similar to a person writing a summary from their thoughts.
Currently, there are machine learning-based summarization tools available. However, finding language-specific models poses a significant challenge. Although significant efforts have been made in Bengali extractive summarization, there are few instances of Bengali abstractive summarization\cite{amin2024sentiment}. Most existing works rely on basic machine learning techniques and face limitations such as small datasets. The absence of established datasets has impeded significant progress in encoder-decoder based summarization systems. Therefore, the most difficult aspect for Bengali Abstractive Summarization (BANS) is the preparation of a well-organized and sanitized dataset.

\section{Related Works}

Numerous methodologies for abstractive text summarization have been identified. Yeasmin et al. \cite{yeasmin2017study} have provided a comprehensive review of various abstractive techniques. Consequently, Our focus was directed towards our research regarding methods for generating abstractive text summaries specifically for the Bengali language.
Haque et al. (2020) provide a comprehensive overview of 14 methods for summarizing Bengali text encompass both extractive and abstractive approaches\cite{haque2020approaches}. In 2004, Islam et al. \cite{islam2004bhasa}  were trailblazers in Bengali extractive summarization, using document indexing combined with keyword-based information retrieval methods. Following this, Uddin et al. \cite{uddin2007study} adapted techniques from Extractive text summarization for Bengali texts using English as the intermediate language. In 2010, Das et al. \cite{das2010topic}
Das et al. (2010) utilized methods such as theme identification and PageRank algorithms for extractive summarization. Kamal Sarkar first proposed Bengali extractive summarization based on sentence ranking and stemming processes \cite{sarkar2012bengali}, with further improvements by Efat et al. \cite{efat2013automated}. and Haque et al. suggested an extractive approach based on key phrases \cite{haque2016enhancement} and a method for ranking sentences through pronoun replacement \cite{haque2017innovative}. In 2017, several notable techniques for Bengali  \cite{10561739}extractive summarization emerged, including a Abujar et al. \cite{abujar2017heuristic}proposed a heuristic approach, Akther et al. \cite{akter2017extractive}employed a K-means clustering method, and Chowdhury et al. utilized the Latent Semantic Analysis (LSA) technique. \cite{chowdhury2017approach}. Ghosh et al. \cite{ghosh2018rule} were the first to use a graph-based sentence scoring feature for Bengali summarization. Additionally, Sarkar et al. \cite{sarkar2018automatic} and Ullah et al. \cite{ullah2019opinion} have put forward extractive approaches based on term frequency and cosine similarity, respectively. Recently, Munzir et al. \cite{al2019text} initiated a deep neural network-based approach for Bengali extractive summarization. Abujar et al. \cite{abujar2019approach} introduced Word2Vec-based word embedding for Bengali text summarization. Talukder et al. \cite{talukder2019bengali} then proposed an abstractive approach using bi-directional RNNs with LSTM at the encoder and attention mechanisms at the decoder. Our study also employs an LSTM-RNN based attention model similar to Talukder et al. \cite{talukder2019bengali}, but we apply attention to both the encoder and decoder layers and conduct a comparative study with the corresponding results and datasets. Additionally, Abujar et al. \cite{abujar2020bengali} 
introduced an LSTM-RNN-based text generation method for creating abstractive text summaries in Bengali \cite{10561876}.
Our system utilizes the concepts Lopyrev et al.  \cite{lopyrev2015generating}discussed this in their work. We adopt the seq2seq model and LSTM encoder-decoder architecture based on the frameworks introduced by Sutskever et. al.\cite{sutskever2014sequence}  and Bahdanau et. al.  \cite{bahdanau2014neural}, respectively. Additionally, the attention techniques for the components comprising the encoder and decoder based on the concepts from Luong et al. \cite{luong2015effective} and Rush et al.  \cite{rush2015neural} have also explored. We used a special computer program to understand how sentences are built, like Vinyals and his team did.We used a special computer program to understand how sentences are built, like Vinyals and his team did \cite{vinyals2015grammar}.

\section{Methodology}
This section delineates the methodology proposed for a Bengali news summarizer employing a neural network-based approach.

\subsection{DataSet}
The presence of a standardized dataset is crucial for enabling effective text summarization.

\begin{figure}[ht]
   \includegraphics[width=0.5\textwidth]{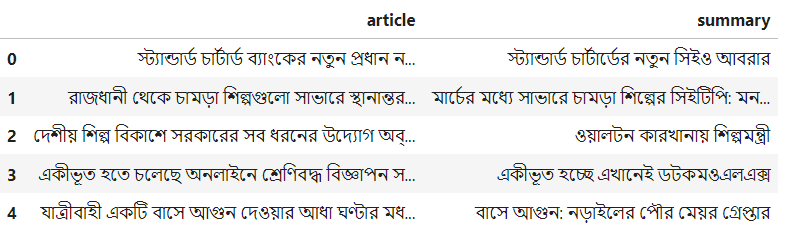}
    \caption{Dataset Overview}
    \label{fig:datasetoverview}
\end{figure}

Inspired by Hermann et al. \cite{hermann2015teaching} and drawing insights from established English datasets such as the CNN-Daily Mail dataset, we conceived the idea of creating a standardized dataset\cite{roy2024english}. Given the scarcity of publicly available datasets for Bengali summarization, our focus turned to the online news portal bangla.bdnews24.com, which offered a repository of news articles and corresponding summaries \cite{islam2024enhancing}. Employing a custom crawler, we collected 19,352 articles and summaries spanning various domains including sports, politics, and economics. Online news sources often feature extraneous content such as advertisements, non-Bengali words, and hyperlinks. To mitigate this, we developed a data cleaning program to systematically filter out irrelevant information from our dataset.

\begin{table}[ht]
    \centering
    \caption{Statistics of the Dataset}
    \label{table:MLModelAcc}
    \begin{tabular}{|c|c|}
    \hline
    \textbf{Total Articles} & 19,096 \\ [1.5ex]
    \hline
    \textbf{Number of Summaries} & 19,096 \\ [1.5ex]
    \hline
    \textbf{Maximum words in an article} & 76 \\ [1.5ex]
    \hline
    \textbf{Maximum words in a summary} & 12 \\ [1.5ex]
    \hline
    \textbf{Minimum words in an article} & 5 \\ [1.5ex]
    \hline
    \textbf{Minimum words in a summary} & 3 \\ [1.5ex]
    \hline
    \end{tabular}
\end{table}

\subsection{Pre-processing and cleaning}
Preprocessing steps are applied to cleanse our dataset, underscoring the pivotal role of cleaning in enhancing outcomes before model training. These steps encompass the removal of extraneous spaces, unnecessary characters, and similar adjustments. Subsequently, start and end tokens are incorporated to delineate the beginning and conclusion of sentences.

\subsection{Model Architecture}
Inspired by the impressive results achieved by the LSTM encoder-decoder with attention mechanism as reported in the study by Lopyrev et al. [20], we have developed a similar neural attention model architecture for our research. Our model comprises both LSTM Encoder and LSTM Decoder components, each integrated with attention mechanisms. TensorFlow's embedding layer, embedding\_attention\_seq2seq was employed to numerically represent words for input into the encoders. Following the generation of the decoder’s output, a comparison between the actual and predicted summaries was conducted using the softmax loss function, initiating back-propagation to minimize the loss. Subsequently, a summary was produced with minimized loss. This entire process operates under a seq2seq approach, visually depicted in figure 1.

Let's outline the key elements of our model. Initially, the input sequence is converted into numerical representations through a word embedding layer and then fed into the LSTM encoder in reverse order, as demonstrated by Sutskever et al. [21] suggested this approach to enhance the proximity of the initial words in both the input and output sequences, thereby capturing short-term dependencies more effectively.

Additionally, we employed a greedy LSTM decoder as opposed to a beam search decoder. The output from the encoder initiates the first decoder cell, and then the output of each decoder the cell is input into the subsequent cell. This process integrates attention mechanisms and information from previous decoder cells until the sequence is fully decoded.

\begin{figure}
    \centering
    \includegraphics[width=0.5\textwidth]{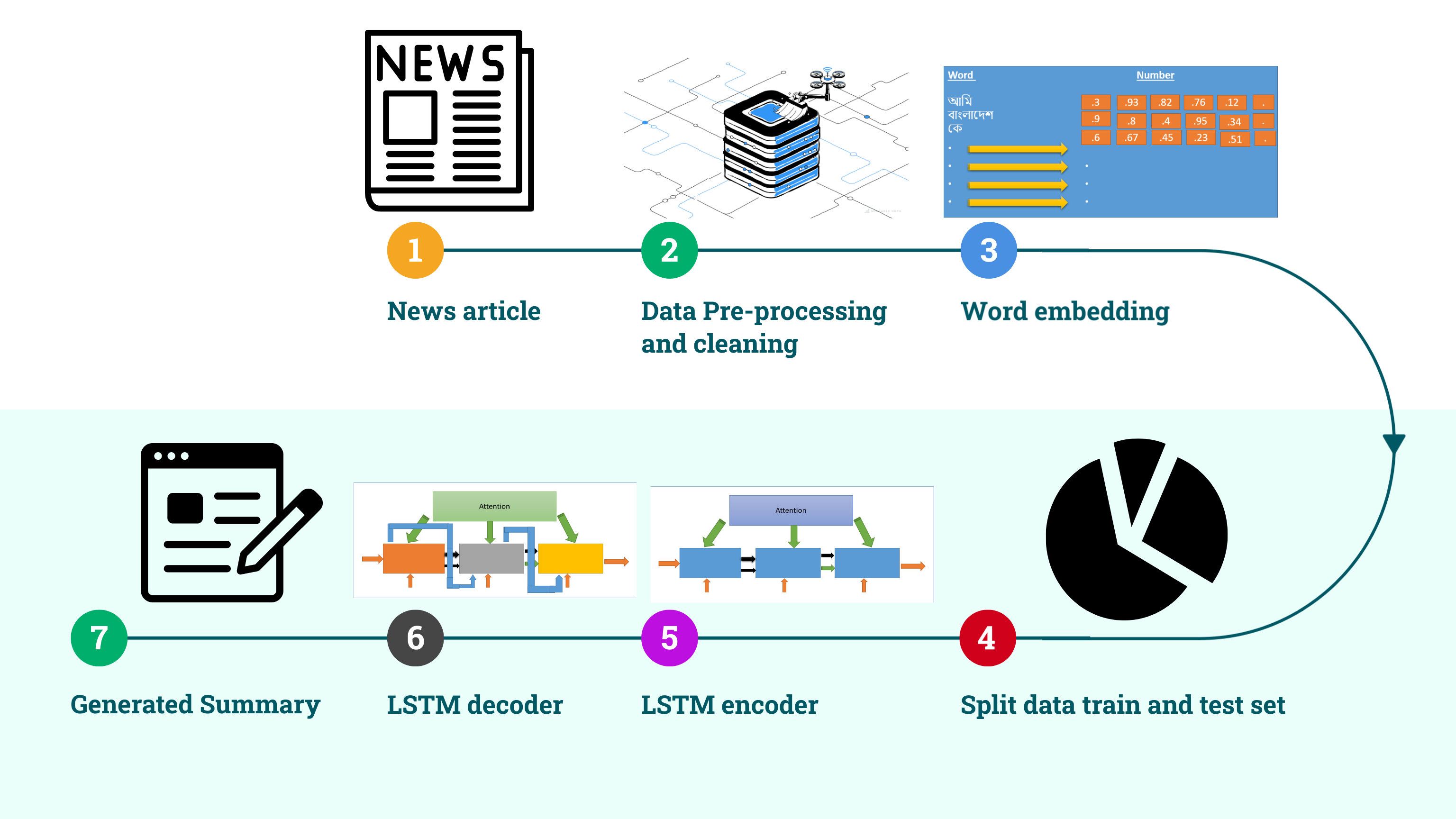}
    \caption{Design of the LSTM-based encoder-decoder model for summarizing Bengali text.}
    \label{fig:methodology}
\end{figure}

\section{Result and Discussion}

\subsection{Evaluation Metrix}

We begin by detailing the dataset employed to evaluate our methods and benchmark them against existing models. Next, we outline the experimental settings used in our research. Finally, we present the performance outcomes, which are assessed utilizing metrics like accuracy, F1-score recall or precision.

\textbf{Accuracy:} This metric is crucial for evaluating the performance of a classifier, indicating the proportion of correct predictions out of the total predictions made. The formula for calculating accuracy is provided in Equation-\ref{eq:1}.
\begin{equation}
\label{eq:1}
\text{Accuracy} = \frac{TP + TN}{TP + TN + FP + FN}
\end{equation} \

\textbf{Precision:} This metric measures the number of true positive predictions made by the model relative to the total number of positive predictions. The formula for calculating precision is shown in Equation-\ref{eq:2}.
\begin{equation}
\label{eq:2}
\text{Precision} = \frac{TP}{TP + FP}
\end{equation} \

\textbf{Recall:} This metric evaluates the model's ability to correctly identify positive instances out of all actual positive instances in the dataset. The formula for calculating recall is defined in Equation-\ref{eq:3}.
\begin{equation}
\label{eq:3}
\text{Recall} = \frac{TP}{TP + FN}
\end{equation} \

\textbf{F1-Score:} This metric assesses the balance between recall and precision. The F1-score reaches its maximum value when recall equals precision. It can be computed using Equation-\ref{eq:4}.
\begin{equation}
\label{eq:4}
\text{F1-Score} = \frac{2 \times \text{Precision} \times \text{Recall}}{\text{Precision} + \text{Recall}}
\end{equation} \

\begin{figure}
    \centering
    \includegraphics[width=.5\textwidth, height=0.3\textwidth]{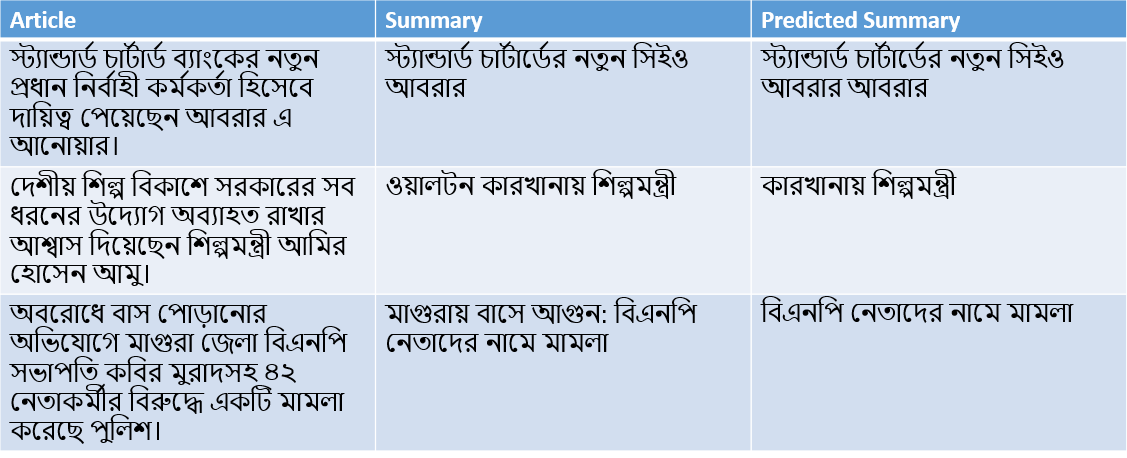}
    \caption{Visual comparison of the generated summary with the predicted summary.}
    \label{fig:predictedsummary}
\end{figure}

\section{Conclution}

In summary, our principal accomplishment involves the creation of a standardized summarization dataset containing 19,096 Bengali news articles, establishing it as the most extensive publicly accessible dataset in this domain. Additionally, we have introduced an encoder-decoder model utilizing neural attention mechanisms tailored for generating abstractive summaries of Bengali news articles, capable of generating coherent sentences that encapsulate the essential information from original documents. Furthermore, we conducted a comprehensive experiment to evaluate the effectiveness of our proposed approach, BANS. Qualitative assessment indicates that our system produces more natural outputs compared to existing methods. Despite the significant success of the LSTM-based encoder-decoder model, challenges arise with reduced performance on longer input sequences, leading to issues such as summary repetition and inaccurate factual reproduction. To overcome these challenges, our future efforts will focus on developing hierarchical encoder models employing structural attention or pointer-generator architectures, and exploring methods for multi-document summarization.

\bibliographystyle{ieeetr}
\bibliography{reference}

\begin{thebibliography}{10}

\bibitem{amin2024sentiment}
A.~Amin, A.~Sarkar, M.~M. Islam, A.~A. Miazee, M.~R. Islam, and M.~M. Hoque,
  ``Sentiment polarity analysis of bangla food reviews using machine and deep
  learning algorithms,'' in {\em 2024 3rd International Conference on
  Advancement in Electrical and Electronic Engineering (ICAEEE)}, pp.~1--6,
  IEEE, 2024.

\bibitem{yeasmin2017study}
S.~Yeasmin, P.~B. Tumpa, A.~M. Nitu, M.~P. Uddin, E.~Ali, and M.~I. Afjal,
  ``Study of abstractive text summarization techniques,'' {\em American Journal
  of Engineering Research}, vol.~6, no.~8, pp.~253--260, 2017.

\bibitem{haque2020approaches}
M.~M. Haque, S.~Pervin, A.~Hossain, and Z.~Begum, ``Approaches and trends of
  automatic bangla text summarization: challenges and opportunities,'' {\em
  International Journal of Technology Diffusion (IJTD)}, vol.~11, no.~4,
  pp.~67--83, 2020.

\bibitem{islam2004bhasa}
M.~T. Islam and S.~M. Al~Masum, ``Bhasa: A corpus-based information retrieval
  and summariser for bengali text,'' in {\em Proceedings of the 7th
  International Conference on Computer and Information Technology}, 2004.

\bibitem{uddin2007study}
M.~N. Uddin and S.~A. Khan, ``A study on text summarization techniques and
  implement few of them for bangla language,'' in {\em 2007 10th international
  conference on computer and information technology}, pp.~1--4, IEEE, 2007.

\bibitem{das2010topic}
A.~Das and S.~Bandyopadhyay, ``Topic-based bengali opinion summarization,'' in
  {\em Coling 2010: Posters}, pp.~232--240, 2010.

\bibitem{sarkar2012bengali}
K.~Sarkar, ``Bengali text summarization by sentence extraction,'' {\em arXiv
  preprint arXiv:1201.2240}, 2012.

\bibitem{efat2013automated}
M.~I.~A. Efat, M.~Ibrahim, and H.~Kayesh, ``Automated bangla text summarization
  by sentence scoring and ranking,'' in {\em 2013 International Conference on
  Informatics, Electronics and Vision (ICIEV)}, pp.~1--5, IEEE, 2013.

\bibitem{haque2016enhancement}
M.~M. Haque, S.~Pervin, and Z.~Begum, ``Enhancement of keyphrase-based approach
  of automatic bangla text summarization,'' in {\em 2016 IEEE Region 10
  Conference (TENCON)}, pp.~42--46, IEEE, 2016.

\bibitem{haque2017innovative}
M.~M. Haque, S.~Pervin, and Z.~Begum, ``An innovative approach of bangla text
  summarization by introducing pronoun replacement and improved sentence
  ranking,'' {\em Journal of Information Processing Systems}, vol.~13, no.~4,
  pp.~752--777, 2017.

\bibitem{10561739}
M.~R. Islam, A.~Amin, and A.~N. Zereen, ``Enhancing bangla language next word
  prediction and sentence completion through extended rnn with bi-lstm model on
  n-gram language,'' in {\em 2024 3rd International Conference on Advancement
  in Electrical and Electronic Engineering (ICAEEE)}, pp.~1--6, 2024.

\bibitem{abujar2017heuristic}
S.~Abujar, M.~Hasan, M.~Shahin, and S.~A. Hossain, ``A heuristic approach of
  text summarization for bengali documentation,'' in {\em 2017 8th
  International Conference on Computing, Communication and Networking
  Technologies (ICCCNT)}, pp.~1--8, IEEE, 2017.

\bibitem{akter2017extractive}
S.~Akter, A.~S. Asa, M.~P. Uddin, M.~D. Hossain, S.~K. Roy, and M.~I. Afjal,
  ``An extractive text summarization technique for bengali document (s) using
  k-means clustering algorithm,'' in {\em 2017 ieee international conference on
  imaging, vision \& pattern recognition (icivpr)}, pp.~1--6, IEEE, 2017.

\bibitem{chowdhury2017approach}
S.~R. Chowdhury, K.~Sarkar, and S.~Dam, ``An approach to generic bengali text
  summarization using latent semantic analysis,'' in {\em 2017 international
  conference on information technology (ICIT)}, pp.~11--16, IEEE, 2017.

\bibitem{ghosh2018rule}
P.~P. Ghosh, R.~Shahariar, and M.~A.~H. Khan, ``A rule based extractive text
  summarization technique for bangla news documents,'' {\em International
  Journal of Modern Education and Computer Science}, vol.~10, no.~12, p.~44,
  2018.

\bibitem{sarkar2018automatic}
A.~Sarkar and M.~S. Hossen, ``Automatic bangla text summarization using term
  frequency and semantic similarity approach,'' in {\em 2018 21st International
  Conference of Computer and Information Technology (ICCIT)}, pp.~1--6, IEEE,
  2018.

\bibitem{ullah2019opinion}
S.~Ullah, S.~Hossain, and K.~A. Hasan, ``Opinion summarization of bangla texts
  using cosine simillarity based graph ranking and relevance based approach,''
  in {\em 2019 International Conference on Bangla Speech and Language
  Processing (ICBSLP)}, pp.~1--6, IEEE, 2019.

\bibitem{al2019text}
A.~Al~Munzir, M.~L. Rahman, S.~Abujar, S.~A. Hossain, {\em et~al.}, ``Text
  analysis for bengali text summarization using deep learning,'' in {\em 2019
  10th International Conference on Computing, Communication and Networking
  Technologies (ICCCNT)}, pp.~1--6, IEEE, 2019.

\bibitem{abujar2019approach}
S.~Abujar, A.~K.~M. Masum, M.~Mohibullah, S.~A. Hossain, {\em et~al.}, ``An
  approach for bengali text summarization using word2vector,'' in {\em 2019
  10th International Conference on Computing, Communication and Networking
  Technologies (ICCCNT)}, pp.~1--5, IEEE, 2019.

\bibitem{talukder2019bengali}
M.~A.~I. Talukder, S.~Abujar, A.~K.~M. Masum, F.~Faisal, and S.~A. Hossain,
  ``Bengali abstractive text summarization using sequence to sequence rnns,''
  in {\em 2019 10th International Conference on Computing, Communication and
  Networking Technologies (ICCCNT)}, pp.~1--5, IEEE, 2019.

\bibitem{abujar2020bengali}
S.~Abujar, A.~K.~M. Masum, M.~Sanzidul~Islam, F.~Faisal, and S.~A. Hossain, ``A
  bengali text generation approach in context of abstractive text summarization
  using rnn,'' {\em Innovations in Computer Science and Engineering:
  Proceedings of 7th ICICSE}, pp.~509--518, 2020.

\bibitem{10561876}
A.~Amin, A.~Sarkar, M.~M. Islam, A.~A. Miazee, M.~R. Islam, and M.~M. Hoque,
  ``Sentiment polarity analysis of bangla food reviews using machine and deep
  learning algorithms,'' in {\em 2024 3rd International Conference on
  Advancement in Electrical and Electronic Engineering (ICAEEE)}, pp.~1--6,
  2024.

\bibitem{lopyrev2015generating}
K.~Lopyrev, ``Generating news headlines with recurrent neural networks,'' {\em
  arXiv preprint arXiv:1512.01712}, 2015.

\bibitem{sutskever2014sequence}
I.~Sutskever, O.~Vinyals, and Q.~V. Le, ``Sequence to sequence learning with
  neural networks,'' {\em Advances in neural information processing systems},
  vol.~27, 2014.

\bibitem{bahdanau2014neural}
D.~Bahdanau, K.~Cho, and Y.~Bengio, ``Neural machine translation by jointly
  learning to align and translate,'' {\em arXiv preprint arXiv:1409.0473},
  2014.

\bibitem{luong2015effective}
M.-T. Luong, H.~Pham, and C.~D. Manning, ``Effective approaches to
  attention-based neural machine translation,'' {\em arXiv preprint
  arXiv:1508.04025}, 2015.

\bibitem{rush2015neural}
A.~M. Rush, S.~Chopra, and J.~Weston, ``A neural attention model for
  abstractive sentence summarization,'' {\em arXiv preprint arXiv:1509.00685},
  2015.

\bibitem{vinyals2015grammar}
O.~Vinyals, {\L}.~Kaiser, T.~Koo, S.~Petrov, I.~Sutskever, and G.~Hinton,
  ``Grammar as a foreign language,'' {\em Advances in neural information
  processing systems}, vol.~28, 2015.

\bibitem{hermann2015teaching}
K.~M. Hermann, T.~Kocisky, E.~Grefenstette, L.~Espeholt, W.~Kay, M.~Suleyman,
  and P.~Blunsom, ``Teaching machines to read and comprehend,'' {\em Advances
  in neural information processing systems}, vol.~28, 2015.

\bibitem{roy2024english}
T.~Roy, M.~R. Islam, A.~A. Miazee, A.~Antara, A.~Amin, and S.~Hossain,
  ``English offensive text detection using cnn based bi-gru model,'' {\em arXiv
  preprint arXiv:2409.15652}, 2024.

\bibitem{islam2024enhancing}
M.~R. Islam, A.~Amin, and A.~N. Zereen, ``Enhancing bangla language next word
  prediction and sentence completion through extended rnn with bi-lstm model on
  n-gram language,'' {\em arXiv preprint arXiv:2405.01873}, 2024.

\end{thebibliography}

\end{document}